\documentclass[10pt,twocolumn,letterpaper]{article}

\usepackage{cvpr}
\usepackage{times}
\usepackage{epsfig}
\usepackage{graphicx}
\usepackage{amsmath}
\usepackage{amssymb}
\usepackage{color}
\usepackage{multirow}
\usepackage{multicol}
\usepackage{subfigure}

\newcommand{\oldversion}[1]{}

\newcommand{\fig}[1]{Fig.~\ref{fig:#1}}
\newcommand{\secc}[1]{Sec.~\ref{sec:#1}}
\renewcommand{\etal}{\emph{et al.}}
\newcommand{\tab}[1]{Table \ref{tab:#1}}
\newcommand{\afterfigure}{\vspace{-0.23cm}}

\usepackage[pagebackref=true,breaklinks=true,letterpaper=true,colorlinks,bookmarks=false]{hyperref}

\cvprfinalcopy 

\def\cvprPaperID{85} 

\ifcvprfinal\fi

\begin{document}

\title{\vspace{-1.0em}Sparse, Smart Contours to Represent and Edit Images}

\vspace{-2cm}
\author{Tali Dekel\textsuperscript{{~1}}$\quad$~
	Chuang Gan\textsuperscript{~2}$\quad$~
	Dilip Krishnan\textsuperscript{~1}$\quad$~
	Ce Liu\textsuperscript{~1}$\quad$William T. Freeman\textsuperscript{~1,3} \vspace{.2cm} \\
	\begin{tabular}{ccc}
		\textsuperscript{1} Google Research & \textsuperscript{2} MIT-Watson AI Lab & \textsuperscript{3} MIT-CSAIL \\ 
	\end{tabular}}

\maketitle

\setcounter{footnote}{0}
\begin{figure}[t]
\vspace{-4in}
\centering
\begin{minipage}{1\textwidth}
	\centering
\includegraphics[width=\textwidth]{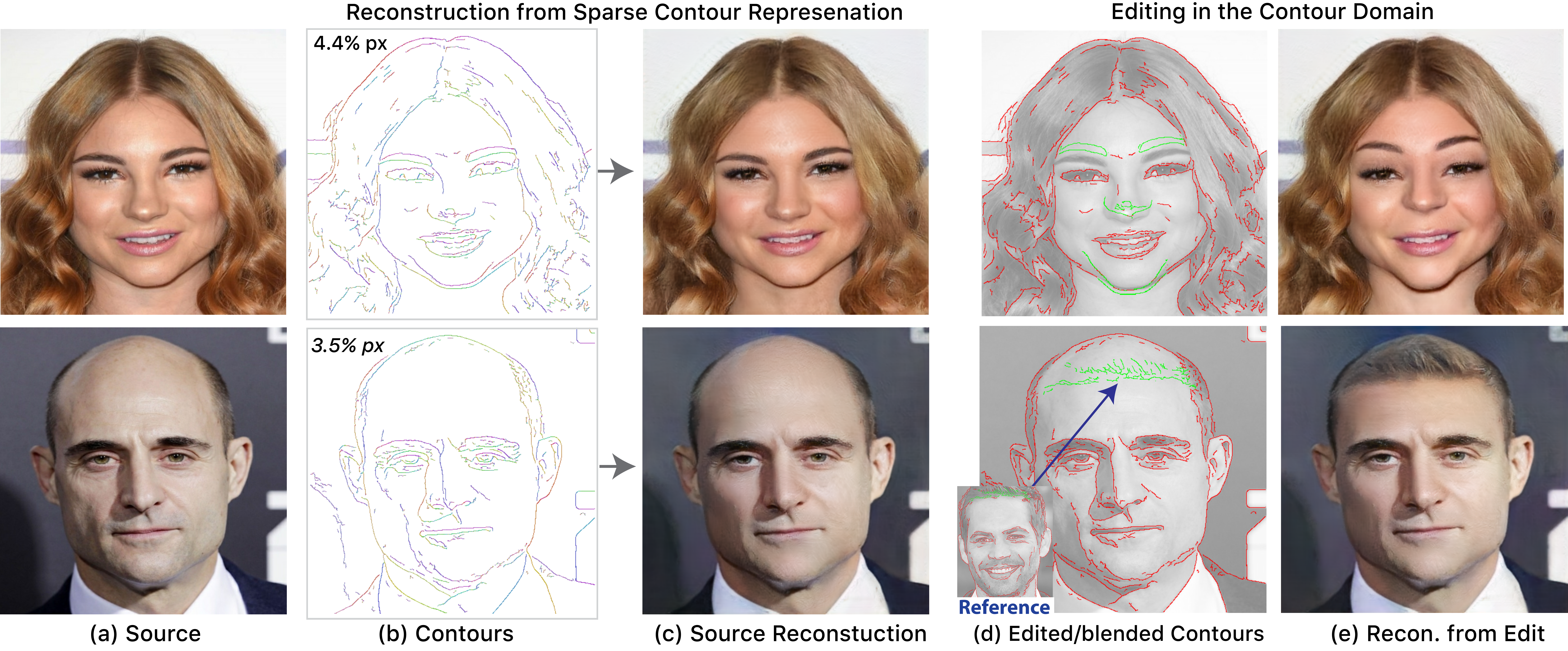}
\caption{Our method produces high quality reconstructions of images from information along a small number of contours: a source ($512\!\times\!512$) image in (a) is reconstructed in (c) from gradient information stored at the set of colored contours in (b)\protect\footnotemark, which are less than 5\% of the pixels. The model synthesizes hair texture, facial lines and shading even in regions where no input information is provided. Our model allows for semantically intuitive editing in the contour domain. Top-right: a caricature-like result (e) is created by moving and scaling some contours in (d). Bottom-right: hairs are synthesized by pasting a set of hair contours copied from a reference image. Edited contours are marked in green while the original contours in red.}
\label{fig:teaser}
\end{minipage}
\vspace{-.15in}
\end{figure} 


\oldversion{
input representations in the form of values at sparse contour locations

reconstructions of images from input representations in the form of values at sparse contour locations: (a) A source image ($512 \times 512$) is reconstructed (c) from gradient information stored at the set of colored edges in (b)\footnotemark. Less than 5\% of pixels are non-zero. The model synthesizes hair texture, facial lines and shading even in regions where no input information is provided. The semantic knowledge encoded in the model and the sparsity of the input allows for intuitive editing in the contour domain. For example, we can easily create caricatures (e) by moving and scaling contour positions (d), or giving the appearance of hair (d) by pasting a set of ``hair'' contours, copied from a reference image shown in the inset; edited contours are marked in green over the source image (d).}

\vspace{1.3em}
\begin{abstract}
\vspace{-0.6em}
We study the problem of reconstructing an image from information stored at contour locations.  We show that high-quality reconstructions with high fidelity to the source image can be obtained from sparse input, e.g., comprising less than $6\%$ of image pixels. This is a significant improvement over existing contour-based reconstruction methods that require much denser input to capture subtle texture information and to ensure image quality. Our model, based on generative adversarial networks, synthesizes texture and details in regions where no input information is provided. The semantic knowledge encoded into our model and the sparsity of the input allows to use contours as an intuitive interface for semantically-aware image manipulation: local edits in contour domain translate to long-range and coherent changes in pixel space. We can perform complex structural changes such as changing facial expression by simple edits of contours. Our experiments demonstrate that humans as well as a face recognition system mostly cannot distinguish between our reconstructions and the source images.

\end{abstract}

\section{Introduction}
\label{sec:introduction}
\setcounter{footnote}{1}\footnotetext{First two principal components of the features is mapped to RGB \cite{baker2011database}.}
\setcounter{footnote}{2}\footnotetext{Project page: \url{https://contour2im.github.io/}}

Contours are a concise and perceptually meaningful representation of the image, as they encode ``things" and not ``stuff" \cite{adelson1991plenoptic}. This makes them appealing for image reconstruction and manipulation. As contours capture shape and object's boundaries, it is desirable to be able to maneuver them (e.g. translating, scaling, copying and pasting) and have the related pixels adapt accordingly, such that the edited images preserve the original image structure and texture details; just as artists use simple sketches to guide drawing sophisticated paintings. Therefore, faithfully reconstructing images from sparse contours, an open question that dates back to the seminal work of David Marr \cite{marr1982computational}, is of great interest as it is the foundation for editing and processing.

\begin{figure}[t!]
    \centering
    \includegraphics[width=0.9\columnwidth]{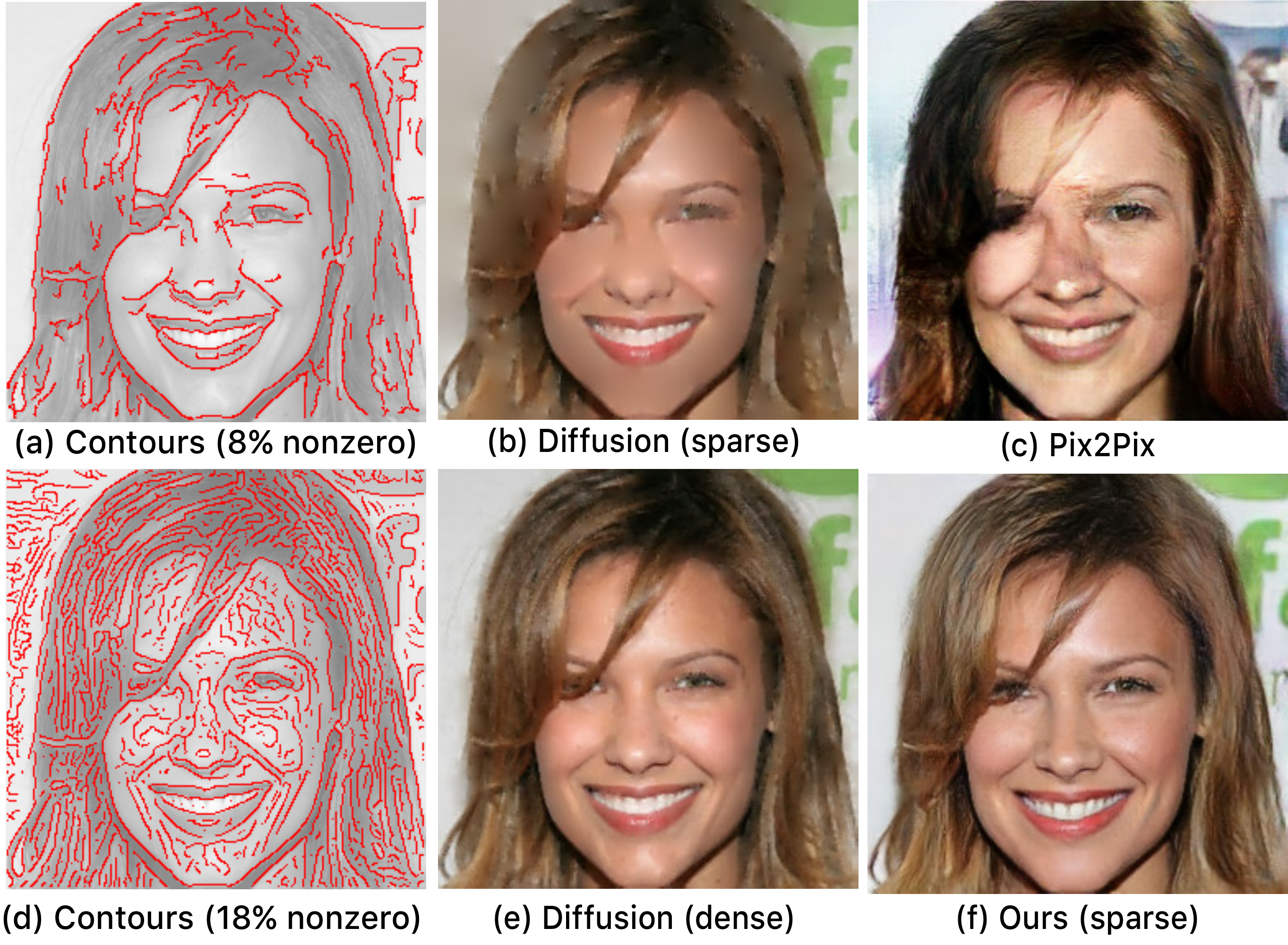}
    \caption{Reconstruction from sparse contours (marked in red in (a)) obtained by: (b) homogeneous diffusion where the input comprises of RGB values sampled at both sides of each edge pixel; (c) Pix2pix  \cite{isola2016image} that takes a binary edge map as input; and (f) our result using gradient information stored at red locations in (a). The source image is shown in Fig.~\ref{fig:pipeline}. (d-e) Dense contours and the corresponding reconstruction obtained by homogeneous diffusion.
    Our approach gives significantly superior reconstruction compared to (b-c) and (e) which is supplied with much denser input.}
    \label{fig:teaser_baselines}
    \afterfigure
\end{figure}

A binary contour map is often insufficient to preserve fidelity for reconstruction (e.g. \cite{isola2016image}, Fig.~\ref{fig:teaser_baselines}(c)). Therefore, local image information such as gradients or color have been combined with contour locations and has been studied extensively in the literature on diffusion-based methods \cite{Elder99,Elder01}.  

However, such diffusion based methods are not applicable for image editing because of their inability to synthesize texture and missing content. High-quality reconstruction often requires dense contours, which precisely forfeits the original purpose of conciseness and ease of manipulation (see \fig{teaser_baselines}(d-e)). When the contours are sparse, the reconstruction loses important image details such as texture (see \fig{teaser_baselines}(a-b)). 

In this paper, we propose a new method, based on deep generative models, to resolve the conflict between high \emph{fidelity} and high \emph{sparsity}. Instead of forcing contours to model textures, details and fine structures, our model learns to hallucinate it appropriately, just from a sparse contour representation, even in large regions where no input information is provided (see Fig.~\ref{fig:teaser}(a-c)). Specifically, we assume that the correlation between contours and textures is well encapsulated in a class of images, such as faces, dogs and birds. For instance, knowing that a contour map is of a person's face, our model can fill in the details of hairs and facial expressions based on the statistical correlation trained on a set of facial images. To this end, we develop a cascade of two networks, splitting the overall task into two more tractable problems. The first network reconstructs the overall image structures and colors, while the second network recovers texture and fine details.

Extensive experiments show that with our model, high fidelity image reconstruction can be obtained from information stored at a small fraction of contour pixels, as small as $\sim$5\% for a $512\times512$ image (see Fig.~\ref{fig:teaser}(a-c)). Working with such highly sparse contours, allows us to explore their use for image editing. Our results demonstrate that our models encode semantic information about the training data. Hence, local edits in the contour domain are translated into coherent changes in the pixel space (e.g., dragging the eyebrows of a person up leads to changes in the facial lines that connect the eyebrow to the nose, see Fig.~\ref{fig:teaser}(d-e)). We show various image editing examples such as creating caricatures, changing facial expression or generating hair or fur texture.

\section{Related Work}
\label{sec:related}
We briefly survey relevant image editing and reconstruction literature. Elder \cite{Elder99, Elder01} explored the completeness of contour representations, and their use in image editing tasks using diffusion based methods. Diffusion curves are vector-based primitives that have been suggested for creating smooth, shaded images \cite{orzan2013diffusion}. Similar representations have been explored for compression of piece-wise smooth images such as depth maps or cartoon images \cite{mainberger2011edge}. 


In contrast to the above mentioned methods, a number of exemplar-based approaches for image editing have been proposed. Prominent among these are patch-based methods \cite{barnes2009patchmatch, simakov2008summarizing} and seam carving \cite{avidan2007seam}. These techniques copy patch information to create high-quality edited images. However, they are oblivious to the semantic content of the image, often failing to produce large, coherent changes. Edits often require human intervention in the form of geometric constraints. 

Deep neural network based image synthesis approaches provide an alternative approach for image editing tasks \cite{iizuka2017globally,Gatys_2016_CVPR,dosovitskiy2016generating,pathak2016context,yeh2016semantic}. Many of these works rely on autoencoder architectures and pixel losses, which have difficulty reconstructing textures. Superior results are possible with the use of generative adversarial networks (GANs) \cite{goodfellow2014generative}.



Unconditional GANs synthesize images from a stochastic latent variable. However, fine user control over the synthesized image is problematic. Several methods (e.g., \cite{zhu2016generative, brock2016neural}) attempt to address this by performing image editing through the low-dimensional learned latent space of the generator network in a GAN. The idea is to optimize the latent representation of an image to satisfy user constrains (e.g., shape or color), while not deviating much from the latent representation of the original image. This approach requires solving a complicated optimization problem by back-propagation which is slow. User constraints are taken into account \emph{implicitly} and so control over the generated image is limited. Finally, although various methods are being developed to stabilize GAN training  \cite{mao2016least,arjovsky2017wasserstein}, synthesizing natural scenes from stochastic inputs is an open problem. 

To combat this, methods have been proposed to condition the GAN on other kinds of inputs \cite{nguyen2016plug, ledig2016photo,IizukaSIGGRAPH2017, pathak2016context, mirza2014conditional,isola2016image,xian2017texturegan,sangkloy2016scribbler,yan2016attribute2image, xian2017texturegan}. Isola \etal ~\cite{isola2016image} synthesize images from input label maps or edge maps. They consider only \emph{binary} edges, which leads to low fidelity to the original image. Furthermore, they did not consider the task of image editing but rather focused on image translation from one domain to another. Sangkloy \etal ~\cite{sangkloy2016scribbler} took a step towards more \emph{controllable synthesis} by learning to generate images from input hand-drawn sketches and additional input sparse color strokes. However, their input sketch is much denser than what we consider in this paper, and hence not suitable for complex geometric manipulation. Due to the density of the input, their network does not need to synthesize texture in large regions as in our case. Furthermore, their edits consist of color changes, leaving contours fixed, unlike our edits that modify the contour structure.

\section{Overview}
\label{sec:overview}
\begin{figure}[t!]
    \centering
    \includegraphics[width=\columnwidth]{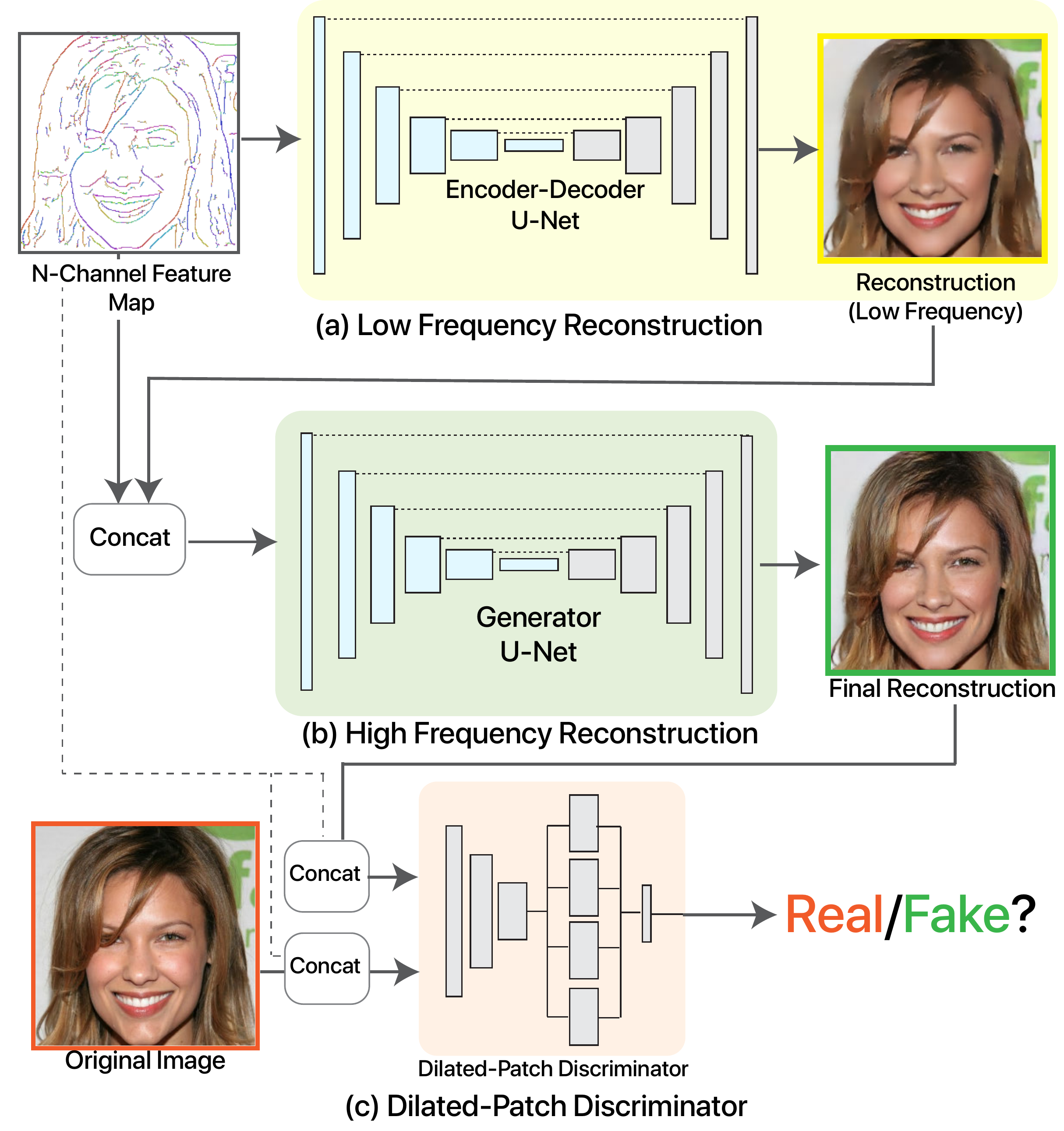}
    \caption{Our model reconstructs an image from a sparse N-channel feature map (typically N=3 or N=6) using a cascade of two ``U-Nets'' \cite{ronneberger2015u}: (a) a Low-Frequency Network (LFN), trained with an $L_1$ pixel loss, that recovers the overall structure and colors of the image; (b) High-Frequency Network (HFN) that is conditioned on the LFN output and the input feature map, produces a textured and detailed reconstruction; the HFN is trained with a combination of pixel loss and adversarial loss. (c) Our conditional discriminator, which incorporates dilated convolutions and aggregation across image patches to better capture high frequencies. ``Concat" refers to concatenating channels of the same spatial resolution along the depth axis.}
    \label{fig:pipeline}
    \afterfigure
\end{figure}

We represent an image by a sparse set of contours (computed using an off-the-shelf edge detector \cite{dollar2013structured}), and an $N$-dimensional feature for each of the contour pixels. In this paper, we have experimented with three types of features: gradients, color and learned features (described further in \secc{input_features}).  
We reconstruct the source image from this input representation using a cascade of two networks, as shown in \fig{pipeline}.

The sparse contour representation is first fed into a network driven by an $L_1$ pixel loss that reconstructs the overall structure and colors of the output image (Fig.~\ref{fig:pipeline}(a)). For example, when training on the VGG face dataset \cite{Parkhi15}, this network recovers face shape, skin tone, hair color and overall shading. We abuse notation slightly and call this network LFN (``Low Frequency Network"), although the reconstruction does contain some high frequency information, given by the input contours. 

The second network takes as input the reconstruction produced by the LFN as well as the original sparse input, and outputs a much more textured and detailed reconstruction of the original image, using an adversarial loss; we call this HFN (``High Frequency Network"). 
Since we work with sparse contours ($\sim6\%$ or less of total image pixels), significant textured regions are not represented in the input. The HFN learns to synthesize plausible texture and structure in these regions. For example, in the case of faces, the hair texture and fine facial lines are synthesized by the HFN, even though very few contours are detected in these regions. 


\begin{figure*}[t!]
    \centering
    \includegraphics[width=\textwidth]{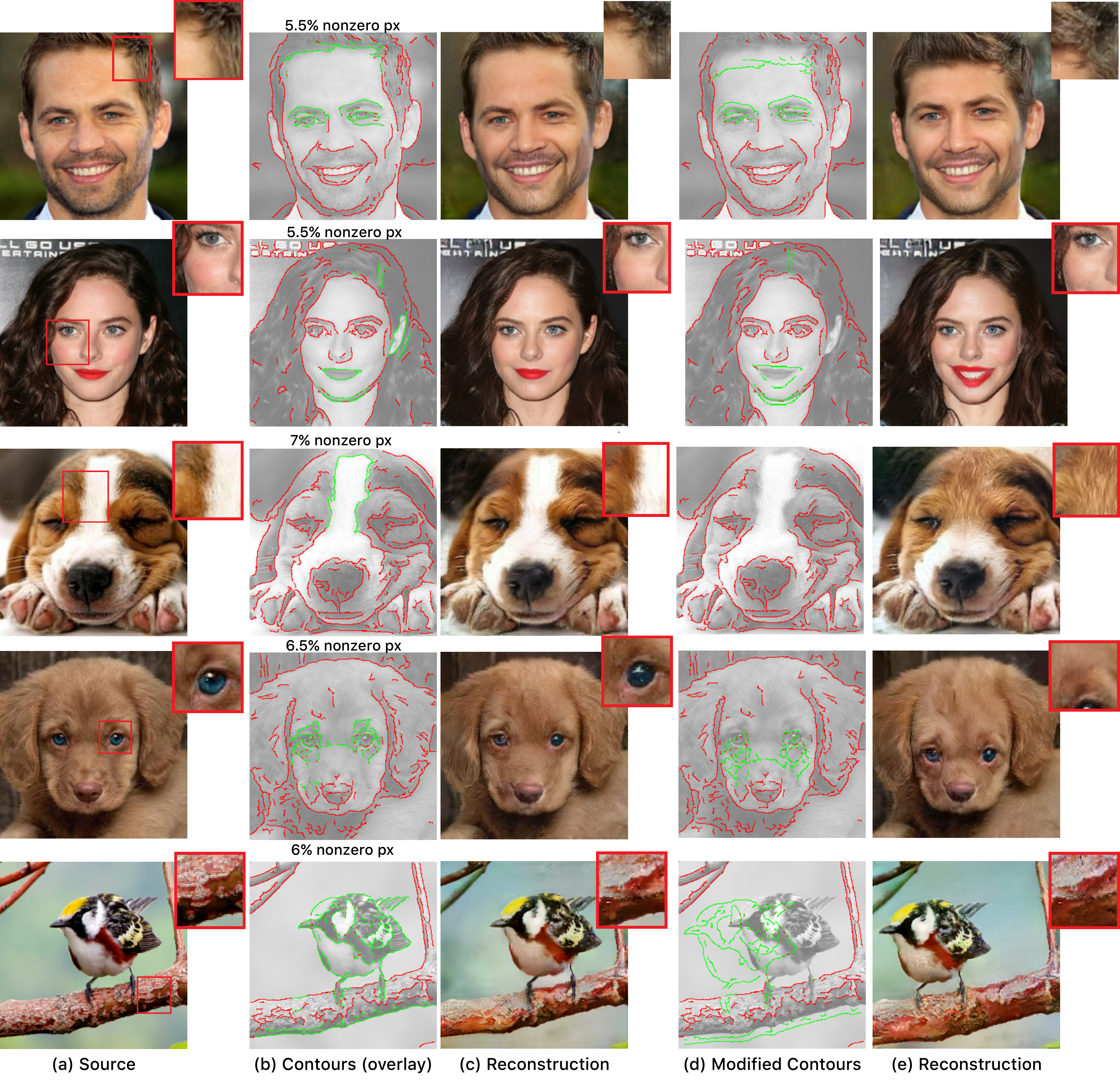}
    \caption{Reconstruction from sparse contours and editing in the contour domain: (a-b) The source images and their input contour maps (colored over the source), respectively. (c) Our final reconstructions using gradient information stored at colored (red and green) contours in (b); our model synthesizes plausible textures and recover fine details in regions with missing data. (d) Edited contours (the green contours in (b) are modified while red contours remain the same). (e) Our reconstructions from edited contours (d); our model translates the modified contours into coherent and semantically meaningful edits of the source images.}
    \label{fig:editing_res}
    \afterfigure
\end{figure*}

\section{Sparse Contour Representation}
\label{sec:input_features}
Given a contour map, an important consideration is what information to encode at each contour position. Color and gradients are common choices among diffusion-based methods (e.g., \cite{orzan2013diffusion, mainberger2011edge}), and gradients have been a useful representation for image editing \cite{perez2003poisson}. We therefore consider these two options, as well as a learned feature representation trained end-to-end with our reconstruction network. We define the feature $f(p)$ at each detected contour point $p$ as follows:\\

\noindent{\bf 1.~Color:~} The orientation of the contour map is computed and $\emph{R,G,B}$ values sampled at both sides of the contour for each edge pixel. That is, $f(p)=\{I^c_d, I^c_b\}_{c\in\{R,G,B\}}$, where $I^c_d$ and $I^c_b$ are values on either side of a contour for channel $c$. This results in a 6-value code per contour location. \vspace{0.1cm}

\noindent{\bf 2.~Gradients:} At each contour point, spatial image derivatives are computed for each of the color channels: $f(p)=\{G_x^c, G_y^c\}_{c\in\{R,G,B\}}$, where $G_x^c, G_y^c$ are the $x$ and $y$ derivatives of the image channel $c$. This also results in a 6-value code per contour pixel.\vspace{0.1cm}

\noindent{\bf 3.~Learned Features:} An $N$ channel feature map is learned end-to-end while training the reconstruction network (we found $N=3$ to be a good balance between quality and complexity). We use a multi-scale representation to encode information from a larger neighbourhood around each edge pixel. We use a simple network that consists of a convolutional layer followed by a branch of \emph{dilated convolution} filters with different sampling rates, employing an architecture similar to {\'a}trous spatial pyramid that presented in \cite{chen2016deeplab}. See Supplementary Materials (SM) for more details.

Of the three choices above, we found that multi-scale learned features result in improved quality of reconstruction (see \fig{multi_scale}). However, for the application of image editing, we found gradient features to have the best trade-off between reconstruction accuracy of the original image and quality of image edits. Although gradient features are the most challenging to invert (the network needs to recover the absolute color values), they provide greater flexibility and robustness to image manipulation. For example, the use of gradients allows blending two sets of contours taken from different images, as shown in \fig{teaser}. This aligns with the literature on image editing in the gradient domain, e.g. \cite{perez2003poisson, bhat2010gradientshop}. 

Color features are more restrictive since they impose constraints on the absolute colors of the reconstruction. Learned features encode multi-scale information, hence the representation of one pixel can be highly correlated with the representation of another. Such correlations between features reduces flexibility to image edits (see \fig{photoshop}(a)). Designing learned features which have the beneficial properties of gradients for image editing, while allowing higher-quality reconstructions, is an interesting area for future research.
 
\section{Model}
\label{sec:model}
\begin{figure}[t!]
    \centering
    \includegraphics[width=\columnwidth]{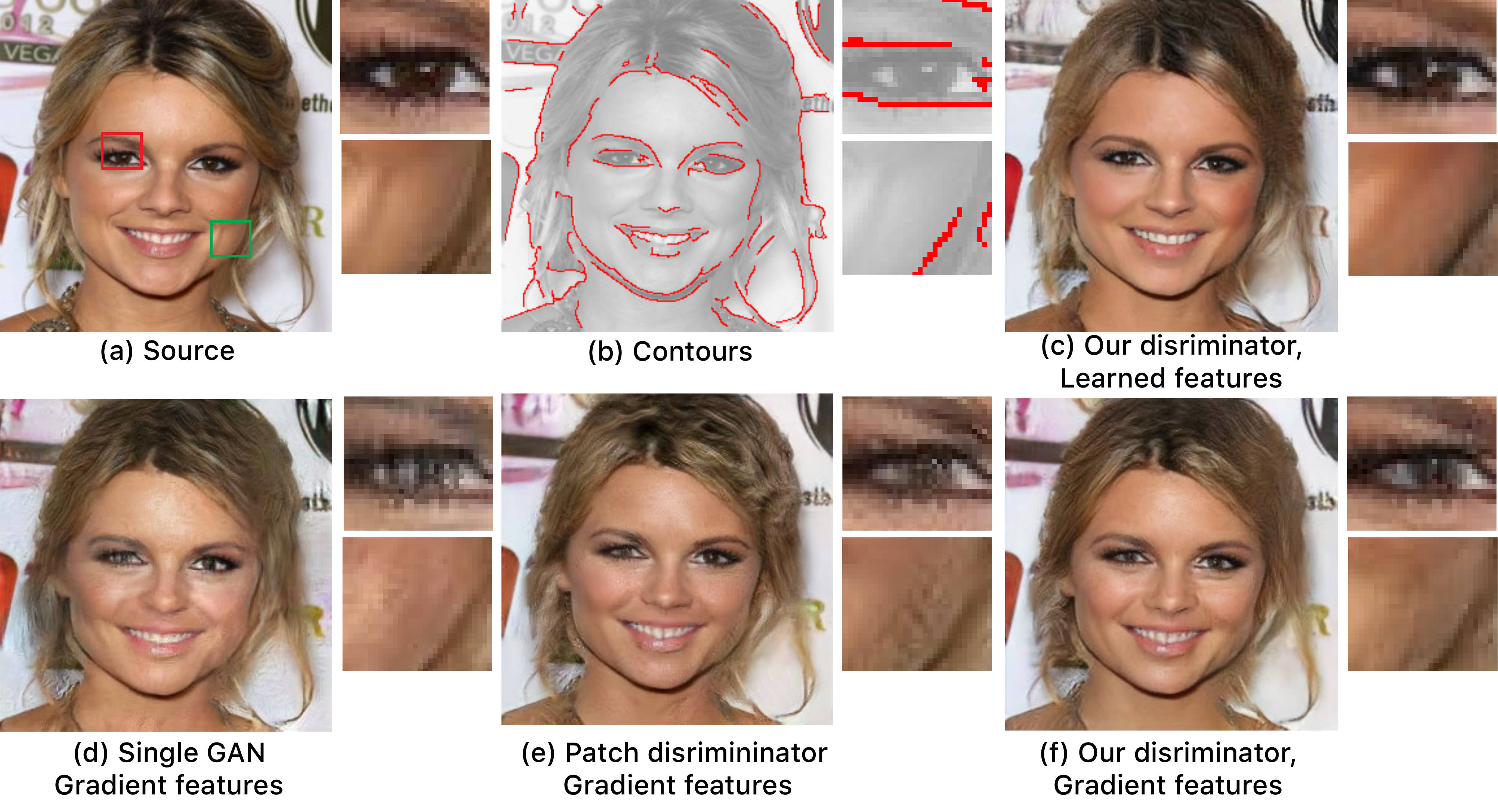}
    \caption{Reconstructions under different network configurations. (a) A source image and (b) the input contour map (marked in red); $5$\% of pixels are non zero. (c) Reconstruction using our cascaded model trained with our dilated-patch discriminator and end-to-end learned features. Second row: reconstructions using gradient features with different models. (d) A single (non-cascaded) GAN. (e) Our cascaded network trained with patch discriminator \cite{isola2016image} and (f) with dilated-patch discriminator. While the learned features are better for reconstruction, they are too restrictive for image editing (see discussion in Sec.~\ref{sec:input_features}).}\afterfigure
    \label{fig:multi_scale}
\end{figure}

As mentioned in \secc{overview} and shown in \fig{pipeline}, our model consists of a cascade of two networks: the first network (LFN) reconstructs the overall structure and colors of the output image from the sparse feature map, whereas the second one (HFN) recovers fine details and texture given the blurry (piecewise smooth) output of the LFN and the original sparse input. This is driven by the training losses. The LFN is trained with an $L_1$ pixel loss between the reconstructed output image and the ground-truth image, which encourages the overall structure and low-frequencies of the output to match the input but is insufficient to reconstruct fine textures and higher frequencies  \cite{johnson2016perceptual}.

The HFN is conditioned on the sparse contours and the output of the LFN, and trained with a combination of $L_1$ pixel loss and an adversarial loss \cite{goodfellow2014generative}. We use a conditional discriminator whose task is to distinguish between the \emph{real} image, and the \emph{fake} output of the HFN (the images are from the same source, i.e. the fake image is reconstructed from the contours of the real image). The weights of the LFN remains fixed during this training. For both the discriminator and generator adversarial losses, we use an $L_2$ loss between the logits of the real and generated samples, following the approach of Mao \etal ~\cite{mao2016least}.

The architecture of the LFN and the generator of HFN is a convolutional encoder and decoder with skip connections between layers of the encoder and decoder \cite{ronneberger2015u}. The architecture of our discriminator (Fig.~\ref{fig:pipeline}(c)) is a combination of a ``patch discriminator'' \cite{isola2016image} and a branch of dilated convolution filters that better capture higher frequencies (\fig{multi_scale})(e-f). See SM for a detailed description.

The network cascade, our patch-dilated discriminator and U-net based decoder, together give high quality reconstructions for both $256 \times 256$ and $512 \times 512$ size images, as demonstrated in Fig.~\ref{fig:teaser} and the supplementary materials. Overall performance is significantly improved over a non-cascaded single GAN (see \fig{multi_scale}), in line with the findings reported in \cite{denton2015deep, zhang2016stackgan}.

\begin{figure*}[t!]
    \centering
    \includegraphics[width=\textwidth]{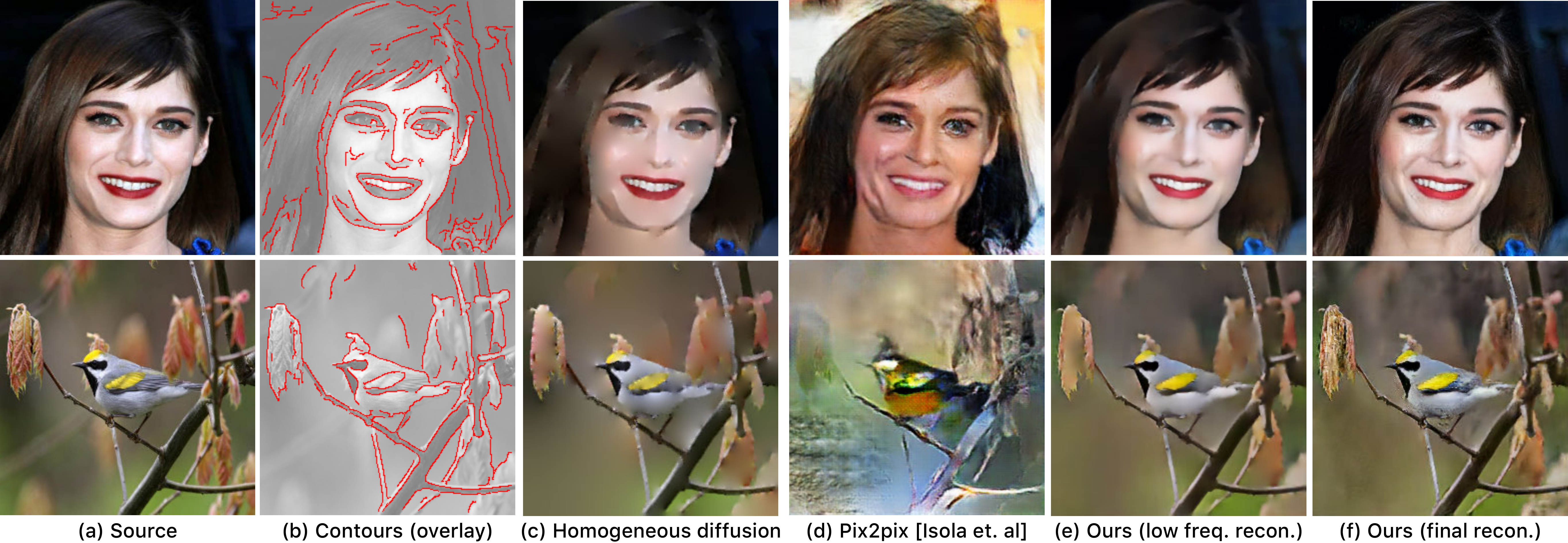}
    \caption{The source image (a) is reconstructed from different representations kept at the same  pixels marked in red (b), using the following methods: (c) Diffusion \cite{Elder01} based solution that propagates RGB values sampled at both sides of each contour pixel. (d) Pix2pix \cite{isola2016image} which uses only binary contours as input. (e) Our LFN output using gradient features stored at each contour pixel and (f) our final HFN output.}
    \label{fig:baselines}
    \afterfigure
\end{figure*}

\newcommand{\myparskip}{\vspace{-7pt}}
\newcommand{\myparagraph}[1]{\vspace{-9pt}\paragraph{#1}}

\section{Experiments}
\label{sec:experiments}
We trained a model for each of the three publicly available datasets from different domains: VGG Faces \cite{Parkhi15}, Caltech-UCSD Birds \cite{WelinderEtal2010} and Stanford Dogs \cite{KhoslaYaoJayadevaprakashFeiFei_FGVC2011}. During training, we  randomly varied the input sparsity for each image (sampled from a Gaussian around 6\% non zero pixels), which allows us to have a wide range of sparsity at inference. See SM for full implementation details. We performed the following experiments to measure the quality of our reconstructions and editing. 

\begin{figure}[t!]
    \centering
    \includegraphics[width=0.5\textwidth]{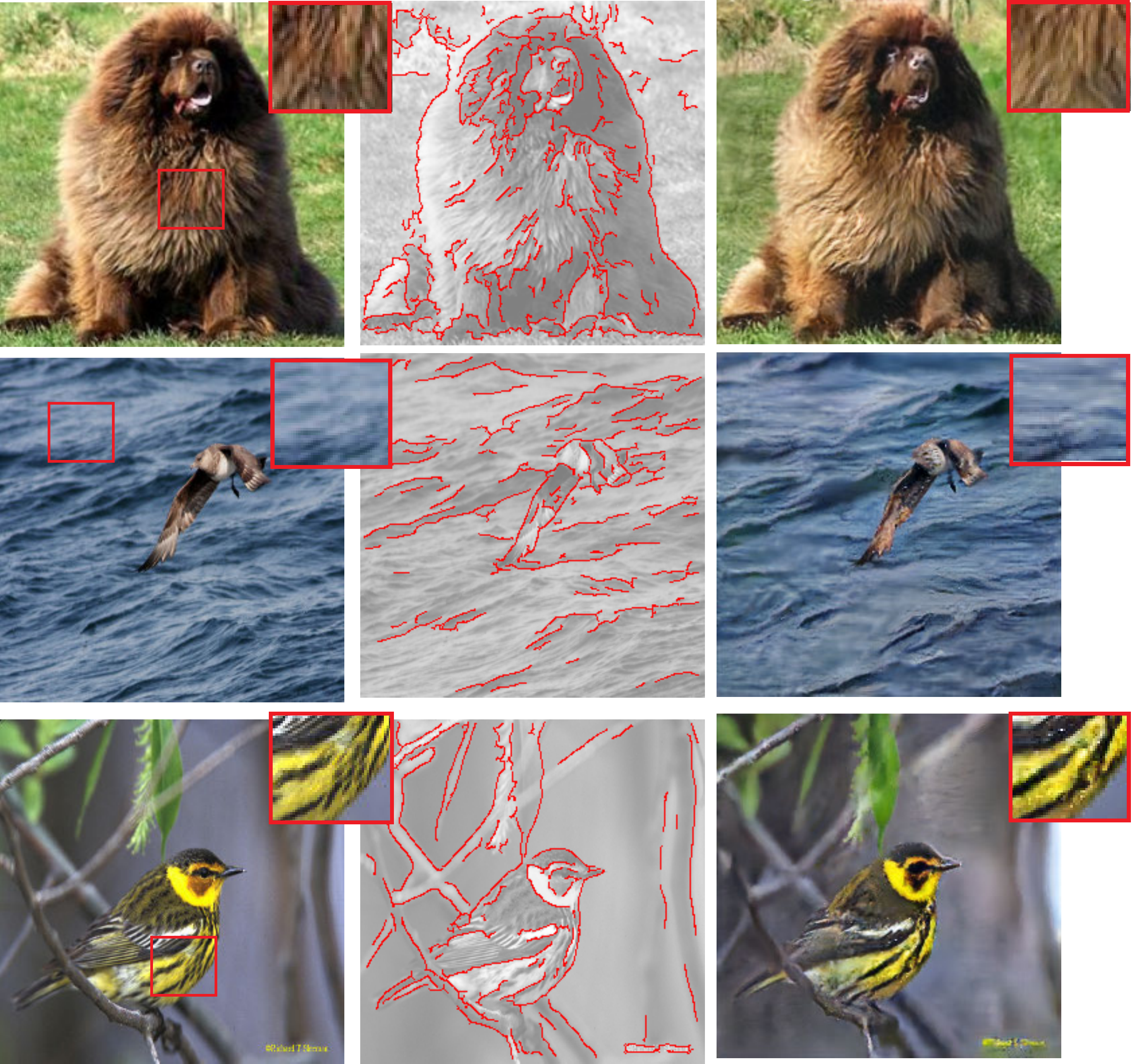}
    \caption{Reconstruction examples of test images from Stanford dogs and Caltech-UCSD datasets. Our network learns to synthesize the textures of the objects e.g., the fur of a dog, as well as the natural backgrounds in the scenes such as water and grass.}
    \label{fig:recon}
    \afterfigure
\end{figure}

\myparagraph{Reconstruction:} \fig{teaser}, Fig.~\ref{fig:teaser_baselines}, \fig{editing_res}(a-c), Fig.~\ref{fig:baselines} and \fig{recon} show sampled reconstruction results obtained by our models when supplied with input gradient features at each contour location (Sec.~\ref{sec:input_features}). As can be seen, our models produce high quality reconstructions w.r.t. the source images from very sparse inputs (4\%-7\% non zero pixels), and recover long range information and details that were not presented by the input. For faces, for example, our model synthesizes hair texture and recover fine details in the eyes, teeth and facial lines.  In \fig{recon}, the texture of the foreground object (fur of a dog, feathers of a bird) is synthesized as well as the texture of the background such as ripples of water, grass, or wood and tree texture. 

\myparagraph{Editing:}  \fig{teaser} and \fig{editing_res}(d-e) show a number of editing results, where the the manipulated contours are marked in green. 
In \fig{teaser} (top row), a caricature of a person is created by moving and scaling sets of contours (with gradient information being transferred as edges are moved around): the woman's eyebrows and nose are moved up and the shape of her jaw is changed. The reconstruction gracefully accommodates these modifications, e.g., by adjusting the facial lines that connect to the nose. The bottom row shows an example of generating plausible hair texture by blending in the contour domain: a set of ``hair contours'' (and their underlying features) is copied from a reference image onto the target image. Our model inpaints the region that was originally bald. This effect of hair synthesis is also shown second row in \fig{editing_res} where contours at the forehead boundary were dragged down. 

\begin{table}[b!]
\vspace{-0.1cm}
    \centering
    \begin{tabular}{|c|c|c|}
         \hline 
         Dataset & \multicolumn{2}{c|}{\%Turkers Labeled Real} \\
         & 1 second & 5 seconds \\
         \hline 
         VGG 256x256 \cite{Parkhi15} & 49.3 & 44.7 \\
         \hline 
         VGG 512x512 \cite{Parkhi15} & 47.2 & 43.5 \\
         \hline 
         Stanford Dogs 256x256 \cite{KhoslaYaoJayadevaprakashFeiFei_FGVC2011} & 48.1 & 46.1 \\
         \hline 
         CUB Birds 256x256 \cite{WelinderEtal2010} & 49.9 & 45.8 \\
         \hline 
    \end{tabular} 
    \vspace{1mm}
    \caption{AMT ``real vs fake" test on different datasets. We report the fraction of generated images (our final reconstructions) that were considered real by the workers, when the real-fake pairs were presented for 1 second or 5 seconds.}
    \label{tab:amt}
    \afterfigure
\end{table}

The second example in \fig{editing_res} shows the creation of a smile by moving and scaling contours. Note the fine facial lines that are generated to accommodate the new expression. Additional edits here include moving the hairline to the center of the head, and inpainting the ear region by erasing their set of contours. 

In the third example, we erased the contours that correspond to the white marking on the dog's forehead (third row). The reconstruction (using a model trained on dogs) depicts seamless color transition and convincing fur-like texture. In the following example, we moved the position of the eyes to give the dog puppy-like proportions.

In the bird example (last row of \fig{editing_res}), we made the tree trunk thinner by shifting up the contours that corresponds to its lower boundary. In addition, we relocated the bird in the scene by pasting its contours in the new location, avoiding the need to accurately segment it from the background. The network is robust to missing edges and can reliably inpaint the holes that are generated in the original location of the bird.

These editing examples demonstrate the necessity of working with \emph{sparse} contours: (i) Achieving these editing effects with denser representation (e.g., Fig.~\ref{fig:teaser_baselines}(d)) would have been significantly more challenging and tedious. (ii) The use of sparse contours during training has encouraged the network to learn semantic interpretations of a scene, giving it the ability to synthesize plausible texture and structure. 

\begin{figure*}[t!]
    \centering
    \includegraphics[width=\textwidth]{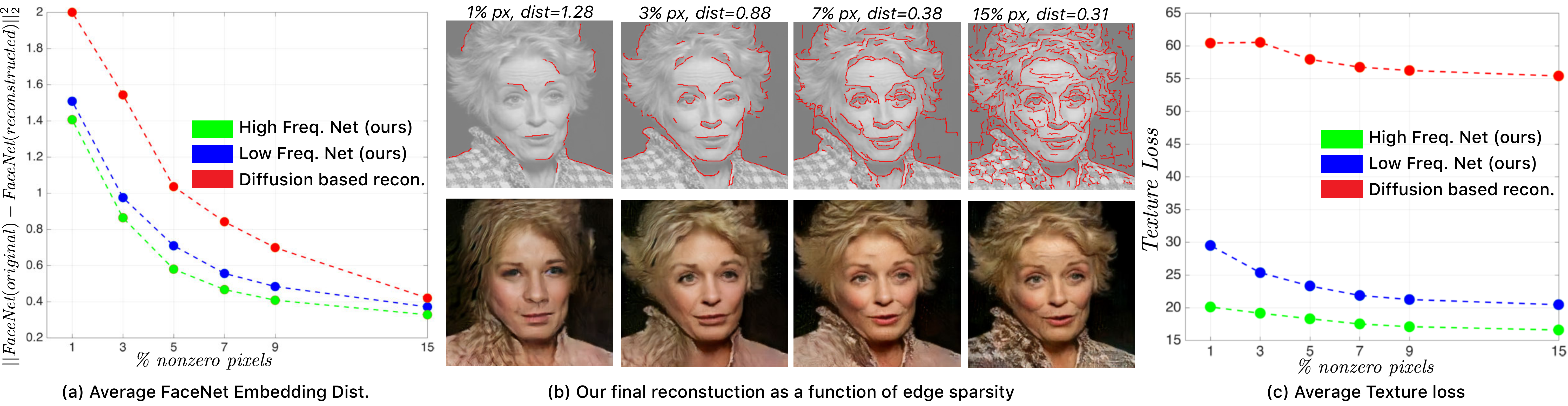}
    \caption{We used FaceNet \cite{schroff2015facenet} to evaluate how close our reconstructions are to the source images in terms of recognition. (a) Average $L_2$ distances between the FaceNet embedding of the source image and its reconstruction, as a function of input contour sparsity. We plot the computed distances (lower is better) for: diffusion-based reconstructions (red), our intermediate reconstructions (LFN, blue) and our final reconstructions (HFN, green). A typical threshold for same/not same classification is $\sim1.2$ \cite{schroff2015facenet}. (b) Our final reconstructions using different sparsity levels (the computed distances shown above each image). (c) Texture loss \cite{Gatys_2016_CVPR} as a function of sparsity; our HFN has the lowest loss, regardless of the sparsity level, which demonstrates the importance of adversarial training. For the images in (b), the texture losses are very similar although the shape changes based on contours.}
    \label{fig:facenet_test}
    \afterfigure
\end{figure*}

\begin{figure}[t!]
    \centering
    \includegraphics[width=0.66\columnwidth]{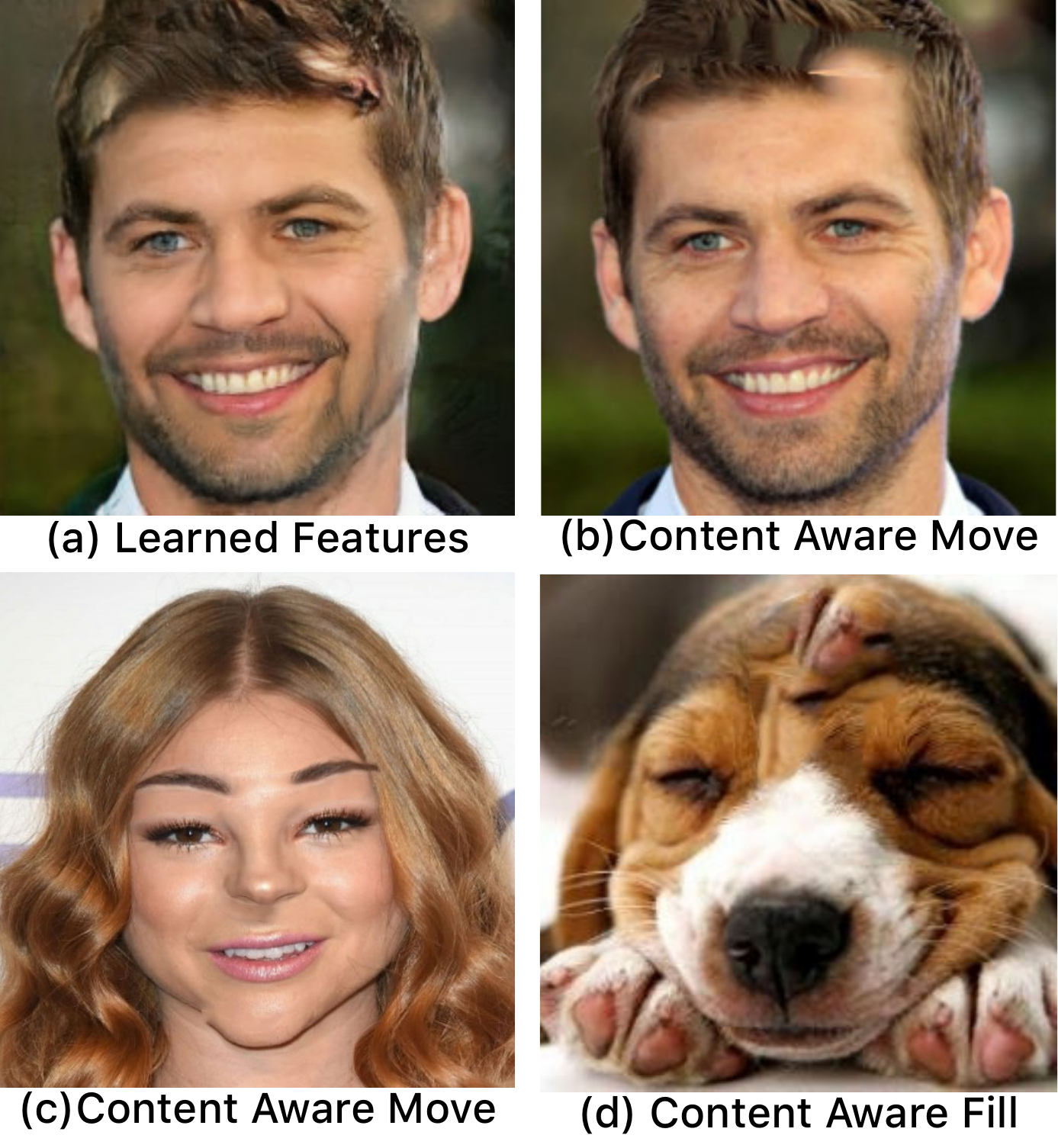}
    \caption{(a) Editing example using our model with learned features. (b-d) Comparison to patch-based Photoshop editing tools: content aware move and content aware fill. Our results using gradients on these images for shown in Fig.~\ref{fig:teaser}(e) and first and third row in Fig.~\ref{fig:editing_res})(e).}
    \label{fig:photoshop}
    \afterfigure
\end{figure}

\myparskip\myparagraph{Model Components and Input Features}
We evaluate the performance of our method using different types of input features as described in Sec.~\ref{sec:input_features}. 
Specifically, we computed the average SSIM \cite{wang2004image} between the source images and the reconstructions, on $100$ randomly sampled images from the VGG, CUB and Dogs datasets. \tab{vgg_ssim} shows the computed scores using either gradient features, color (RGB) or learned features. The learned features, which capture multi-scale information, consistently give the best results, followed by color and then gradients. Note that, in the case of gradients the network needs to recover the unknown absolute color values, which is clearly an ill-posed problem. 
Therefore, in this case the reconstructions  may result in slight color shifts w.r.t. the source images. However, as discussed in Sec.~\ref{sec:input_features}, we found that gradient features provide greater flexibility and robustness to image manipulation. This is exemplified in \fig{photoshop}(a) where the same contour edits, using learned features results in poorer quality than using gradient features (\fig{editing_res} first row); this is due to the spatial correlation between the learned features, which breaks when contours are been manipulated. 

\tab{vgg_ssim} also shows the SSIM scores for a single GAN (using gradient features) vs. our cascaded network. The consistent improvement in the SSIM scores for our model is clear and aligns with other works \cite{zhang2016stackgan,denton2015deep} that demonstrated the benefits of a coarse-to-fine approach. A qualitative example of reconstruction using different network configurations in shown in Fig.~\ref{fig:multi_scale}. 

\begin{table}[b!]
    \centering
    \vspace{-0.3cm}
    \begin{tabular}{|c|c|c|c|c|}
         \hline
         \multirow{2}{*}{Dataset} &  \multirow{2}{*}{GAN} & \multicolumn{3}{c|}{LFN-HFN} \\
          & & Gradients & Color & Learned\\
         \hline
         VGG & 0.786 & 0.812 & 0.859 & 0.867 \\
         \hline
         CUB &  0.697 & 0.740 & 0.763 & 0.783 \\
         \hline
         Dogs &  0.722 & 0.749 & 0.792 & 0.810 \\
         \hline
    \end{tabular}
    \vspace{1mm}
    \caption{Average SSIM scores between 100 randomly sampled images from each test set and their reconstructions using different models.
     From left to right: reconstructions using a single GAN with gradients and reconstructions using our LFN-HFN cascade network trained with gradients, RGB, and learned features. The input contour maps are identical for all models and have approximately 6\% non-zero pixels.}
    \label{tab:vgg_ssim}
    \afterfigure
\end{table}


\myparagraph{Comparison with Baselines}
Fig.~\ref{fig:teaser_baselines} and Fig.~\ref{fig:baselines} show qualitative comparisons to two baselines: (i) homogeneous diffusion--a classical low-level approach for image reconstruction from contour representation (see Sec.~\ref{sec:related}) and (ii) Pix2pix \cite{isola2016image}. 

For diffusion, we follow Elder \cite{Elder01} and use color (RGB) values sampled at either side of a contour location. It is seen that this results in piecewise smooth reconstructions, and fails to recover texture or details at missing contours. This is not surprising since diffusion merely interpolates the color values, without any semantic knowledge. Even when supplied with very dense input (\fig{teaser_baselines}(c)), the diffusion result (d) suffers from spikes and blurriness due to the sensitivity to the location of the constrain and their values. This can also be seen be from the comparison to our low-frequency reconstruction (\fig{baselines}(f)) where the facial highlights and shading are recovered significantly better.

For Pix2pix \cite{isola2016image} that takes as input \emph{binary} edge maps, we trained their network from scratch (using their original PyTorch implementation) on exactly the same data (images and contours) that was used to train our models. Pix2pix, while recovering some texture, fails to recover the scene properties (e.g. skin tone, hair color, bird color) and results in poor quality reconstructions (e.g., tree texture in the background of the the bird image or facial artifacts). This demonstrates the importance of both location and value information. 

\fig{photoshop} shows a comparison to Photoshop's state-of-the-art editing tools, for similar edits as \fig{teaser} and \fig{editing_res}. Since these tools do not have semantic awareness, their results can often be less satisfactory. \vspace{-1mm}

\subsection{Quantitative Evaluation}
Our model synthhesizes high frequency content that may not match exactly the original one. Thus, in the following experiments, we go beyond using standard image similarity measures such as PSNR or SSIM that do not necessarily capture the perceptual quality \cite{johnson2016perceptual}.

\myparagraph{Human Evaluation}
 We evaluated our results using human raters on Amazon Mechanical Turk (AMT), following closely the protocol in \cite{isola2016image}. Workers were presented with pairs of images corresponding to the source image and our reconstruction, and asked to label which one was ``real". Each pair of images was presented for a limited time after which the rater makes their choice. As practice, the first 10 pairs of images were shown without a time limit. We evaluated the ranking of 10 rater, each was given 100 image pairs. Our reconstructions for this test were obtained using gradient based-features at 6\% contour locations. The same test was repeated for 1 and 5 seconds presentation time. The results, reported in \tab{amt}, show that for all datasets, our reconstructed images were hard to distinguish from the real images (a score of $50\%$ would mean perfect confusion between real and fake). As expected. for a 5 second presentation time, it was slightly easier to spot reconstructions.

\myparagraph{Face Recognition Evaluation}
We tested the extent to which our reconstructed faces capture the identity of a person. We used FaceNet \cite{schroff2015facenet}, a well-known face recognition system and measured whether our reconstruction (using different sparsity levels of input) and the source image are classified as the same person. We followed \cite{schroff2015facenet} and computed the squared $L_2$-distance between the $128$-dimensional embedding vectors of the original image and our reconstruction. \fig{facenet_test}(a) shows the average distance over 50 images randomly sampled from the VGG test set when applying our network using gradient input features (as this is the most challenging case for reconstruction) at different contour density levels. We report the error for LFN, HFN and homogenous diffusion. As expected, the performance of HFN is the highest regardless of the contour's sparsity, on average a relative 10\% reduction in error over LFN and nearly 40\% over homogenous diffusion. This shows that reconstructing details is helpful for a face recognition system.

An example of our reconstructions and the input contours are shown in \fig{facenet_test}(b), where the resemblance to the source image gradually increase with density. There is not much loss of information between $15\%$ and $7\%$ nonzero pixels because the network recovers missing high frequency information. Note that even at a sparsity as low as 3\% of pixels, Facenet recognizes the resulting face as being the same as the original (based on the thresholds given in \cite{schroff2015facenet}). \vspace{-2mm}


\myparagraph{Texture/Style Evaluation:} Recent style transfer methods have demonstrated that texture statistics can be captured by the Gram matrix of the activations at some layers in a pretrained recognition network (e.g. \cite{simonyan2014very}). In \cite{Gatys_2016_CVPR}, a \emph{texture-loss} between two images is defined as: $\sum_{l=0}^L w_l \sum_{i,j} (G^l_{ij} - \hat{G}^l_{ij})^2 $, where $l$ denotes the layer, and $G^l_{ij}$ is the inner product between feature map $i$ and $j$ at layer $l$: $G^l_{ij} = \sum_k F^l_{ik} F^l_{jk}$. $G$ and $\hat{G}$ refer to the Gram matrices for two different images. We use this loss (with the same setting as \cite{johnson2016perceptual}) to evaluate the quality of our synthesized texture compared to the source image. Fig.~\ref{fig:facenet_test}(c) shows the computed texture loss for LFN, HFN and diffusion based reconstructions, as a function of the contour's sparsity. The benefit of a GAN loss in synthesizing texture is clearly seen, as the loss of HFN is consistency the lowest and steady over different sparsity levels.

\begin{figure}[t!]
    \centering
    \includegraphics[width=\columnwidth]{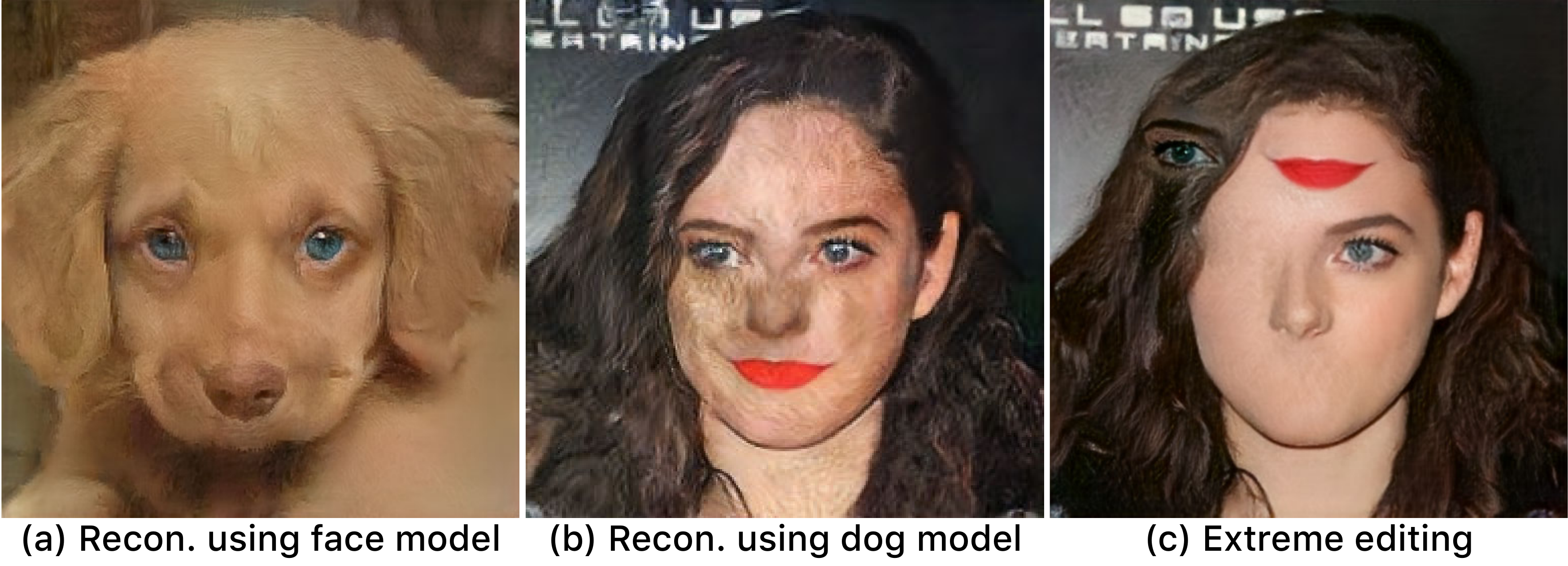}
    \caption{Limitations. (a) A \emph{dog} image is reconstructed using our model when trained on \emph{faces}; (b) shows the reverse. (c) Effect of an extreme edit whereby the result is not semantically meaningful due to the contour constraints.}
    \label{fig:limitation}
    \afterfigure
\end{figure}

\section{Conclusions}
\label{sec:conclusion}
We have presented a deep network model that produces high-quality reconstructions of images, and effective semantically-aware editing, from sparse contour representations. Due to the sparsity and the high level information encoded in our model, the representation has the benefit of being easy to manipulate for large, coherent edits. This is a significant improvement over existing work. 

Our model is domain specific and if the statistics of the source image is very different from the training data, the quality of the reconstruction may deteriorate. For example, applying a model trained on the dogs to a face results in a dog-like appearance (\fig{limitation}(b)), and vice-versa (\fig{limitation}(a)). In both cases, the contours provide a strong constraint on the reconstruction, but the synthesized texture is dominated by the training data. In addition, within the same domain, a distant face (zoomed out) may not be recovered
as well as a portrait shot (as these are the most examples
in the VGG dataset). Finally, some cases (e.g., extreme editing operations) can prevent semantically meaningful reconstructions (\fig{limitation}(c)). 

\newpage
{\small
\bibliographystyle{ieee}
\bibliography{egbib}
}

\end{document}


\maketitle
\section{Implementation Details} 
\label{sec:implementation}
Our models were implemented in Tensorflow. We will release our trained models and code. To achieve the editing effects, we built a simple graphical user interface (GUI) that allows editing the contour map using simple operations such as moving, scaling or erasing sets of contours. Note that although the user edits the contour map, the underlying features change in the appropriate manner (for example, if some contours are removed, then the features associated with those contours are also removed). \afterfigure

\paragraph{Computing Input Representation:} Edge probability maps were extracted using   \cite{dollar2013structured} followed by non-maximum suppression as a post processing. The computed maps were binarized by keeping $x$ percentile of edges, where $x$ is the desired percentage of nonzero pixels in the binary edge map. We group the edges into contours and filter out short contours (length less then 10 pixels). Gradients are computed by simple forward differences. \afterfigure
\paragraph{LFN and HFN:}
Both LFN and HFN are ``U-Nets'' \cite{isola2016image}. We adopt the notations in \cite{isola2016image} and denote a convolution layer with $k$ filters followed by batch normalization by $\texttt{Ck}$ for ReLU activation, and $\texttt{Clk}$ for LeakyReLU (slop 0.2). For 256$\times$256 images the number of filters in the encoder  is given by: $\texttt{Cl64-Cl128-Cl256-Cl512-Cl512-Cl512}$, hence the spatial resolution of the bottleneck is $4\times4$. The number of filters in each layer of the decoder is given by: $\texttt{C512-C512-C256-C128-C64-C64}$ (filters in decoder are bigger because of the skip connections). All convolutional filters are size $4 \times 4$, and we use strides of size $2$.\afterfigure

\paragraph{Dilated-Patch Discriminator}
Our discriminator (Fig.~3(c)) is a combination of a ``patch discriminator'' \cite{isola2016image} and a branch of dilated convolution filters. Let $D_{r}k$ denote a dilated convolution with sampling rate $r$, leaky ReLU and $k$ filters. Then our patch discriminator architecture is given by $\texttt{Cl64-Cl128-Cl256}$ followed by $4$ parallel dilated convolutions ${\{\texttt{D}_{2}\texttt{256}, \texttt{D}_{4}\texttt{256}, \texttt{D}_{8}\texttt{256}, \texttt{D}_{12}\texttt{256}}\}$, which are concatenated, and a final convolution layer with a single channel output.

The network used for learning the input representation (see Sec.~4) is similar and consists of a branch of  dilated convolutions, each followed by regular a convolution layer. That is,
{\small ${\{\texttt{D}_{4}\texttt{32}-\texttt{C3}, \texttt{D}_{8}\texttt{256}-\texttt{C3}, \texttt{D}_{12}\texttt{256}-\texttt{C3}, \texttt{D}_{16}\texttt{256}-\texttt{C3}}\}$}, which are added to form the final output.\afterfigure

\paragraph{Training hyper-parameters and details}
For each dataset, LFN and HFN were trained from scratch (weights were  
initialized from a Gaussian distribution with mean 0 and
standard deviation 0.02). We used Adam optimizer with $\beta=0.5, \epsilon=1e^{-4}$, and batch size of $16$ in all training runs except when training on 512$\times$512 where we used size of $8$. We used learning rate of $0.0002$ for the generator, and $0.00002$ for the discriminator, with decay rate of $0.98$ every $10000$ steps. During training we resized the images such that the small dimension is at the desired resolution and then randomly cropped to desired size. The relative weight of the adversarial loss vs. reconstruction loss was 100 in all our experiments.

During the HFN training, the weights of the LFN remained fixed, and we alternate between stepping the discriminator and generator in ratio of 2:1 in favor of the discriminator. When working with learned features, the weights feature network remained fixed during HFN training.

 We trained on the VGG dataset at two spatial resolutions: $256 \times 256$ and $512 \times 512$. For Birds and Dogs, we used the original train/test splits, and for the VGG dataset we filtered out low resolution images from the train and test sets, to avoid reconstruction of JPEG artifacts. The VGG has 30227 samples and was trained for 210 epochs for both LFN and HFN; CUB Birds has 8855 samples and was trained for 720 epochs; Dogs dataset has 12000 training samples and was trained for 500 epochs.

\paragraph{Texture Loss} We use the texture loss in Section 6.1 to evaluate our reconstructions. This loss was defined in \cite{ustyuzhaninov2016does}.

\bibliographystyle{ieee}
\bibliography{egbib}